\documentclass[sigconf]{acmart}

\usepackage{amsmath}

\usepackage{amssymb}
\usepackage{graphicx}
\usepackage{booktabs}
\usepackage{tabularx}
\usepackage{colortbl}
\usepackage{adjustbox}
\usepackage{makecell}
\usepackage{multirow}
\usepackage{array}
\usepackage{makecell}
\usepackage{xcolor}
\definecolor{darkgold}{rgb}{191,144,0}
\definecolor{lightgray}{gray}{0.9}
\usepackage{threeparttable}
\usepackage{subcaption}
\usepackage{pifont}
\usepackage{enumitem}
\usepackage{algorithm}
\usepackage{algorithmic}
\AtBeginDocument{%
  }

\copyrightyear{2026}
\acmYear{2026}
\setcopyright{cc}
\setcctype{by}
\acmConference[MM '26] {Proceedings of the 34th ACM International Conference on Multimedia}{November 10--14, 2026}{Rio de Janeiro, Brazil.}
\acmBooktitle{Proceedings of the 34th ACM International Conference on Multimedia (MM '26), November 10--14, 2026, Rio de Janeiro, Brazil}
\acmISBN{979-8-4007-2213-4/2026/11}
\acmDOI{10.1145/3767308.3835801}

\begin{document}

\title{HiSC: Hierarchical Spatial Clustering Token Compression for Efficient 3D Scene Understanding}

\author{Jiuhe Qu}
\authornote{Jiuhe Qu and Yingping Liang contributed equally to this work.}
\orcid{0009-0003-6924-6180}
\email{qujiuhe@bit.edu.cn}
\affiliation{%
  \institution{Beijing Institute of Technology}
  \city{Beijing}
  \country{China}
}

\author{Yingping Liang}
\authornotemark[1]
\orcid{0000-0001-5385-0015}
\email{liangyingping@bit.edu.cn}
\affiliation{%
  \institution{Beijing Institute of Technology}
  \city{Beijing}
  \country{China}
}

\author{Ying Fu}
\correspondingauthor
\orcid{0000-0002-6677-694X}
\email{fuying@bit.edu.cn}
\affiliation{%
  \institution{Beijing Institute of Technology}
  \city{Beijing}
  \country{China}
}

\renewcommand{\shortauthors}{Qu et al.}

\begin{abstract}
3D vision-language models (3D VLMs) enable spatial reasoning over multi-view scenes but suffer from substantial token redundancy due to duplicated observations and large uninformative regions, leading to high computational cost. Although visual token compression has shown promise in accelerating 2D VLMs, it fails to capture the structured nature of 3D scenes and leads to incomplete spatial coverage and loss of fine-grained details. In this paper, we propose \textbf{HiSC}, a training-free framework for hierarchical spatial clustering token compression in 3D VLMs. HiSC lifts token compression from token-level selection to cluster-level processing by organizing tokens into spatially grounded clusters using joint geometric and semantic cues. Specifically, we first introduce a \textbf{spatial graph-based merging (SGraM) strategy} that models cross-view redundancy as spatial connectivity and consolidates physically consistent regions, effectively merging extremely similar redundant tokens prior to LLM inference. We then propose a \textbf{spatial clustering-based pruning (SCluP) paradigm} within LLM inference, which performs hierarchical compression across clusters and within clusters, preserving object instance completeness while retaining fine-grained details for important regions. Extensive experiments on diverse 3D reasoning benchmarks show validate the effectiveness of HiSC, particularly under high visual token pruning ratios. Besides, HiSC achieves over 90\% token reduction with minimal performance degradation. Code is accessible at \url{https://github.com/elecreak/HiSC}.
\end{abstract}




\maketitle

\begin{figure}[!t]
\centering
\includegraphics[width=0.97\linewidth]{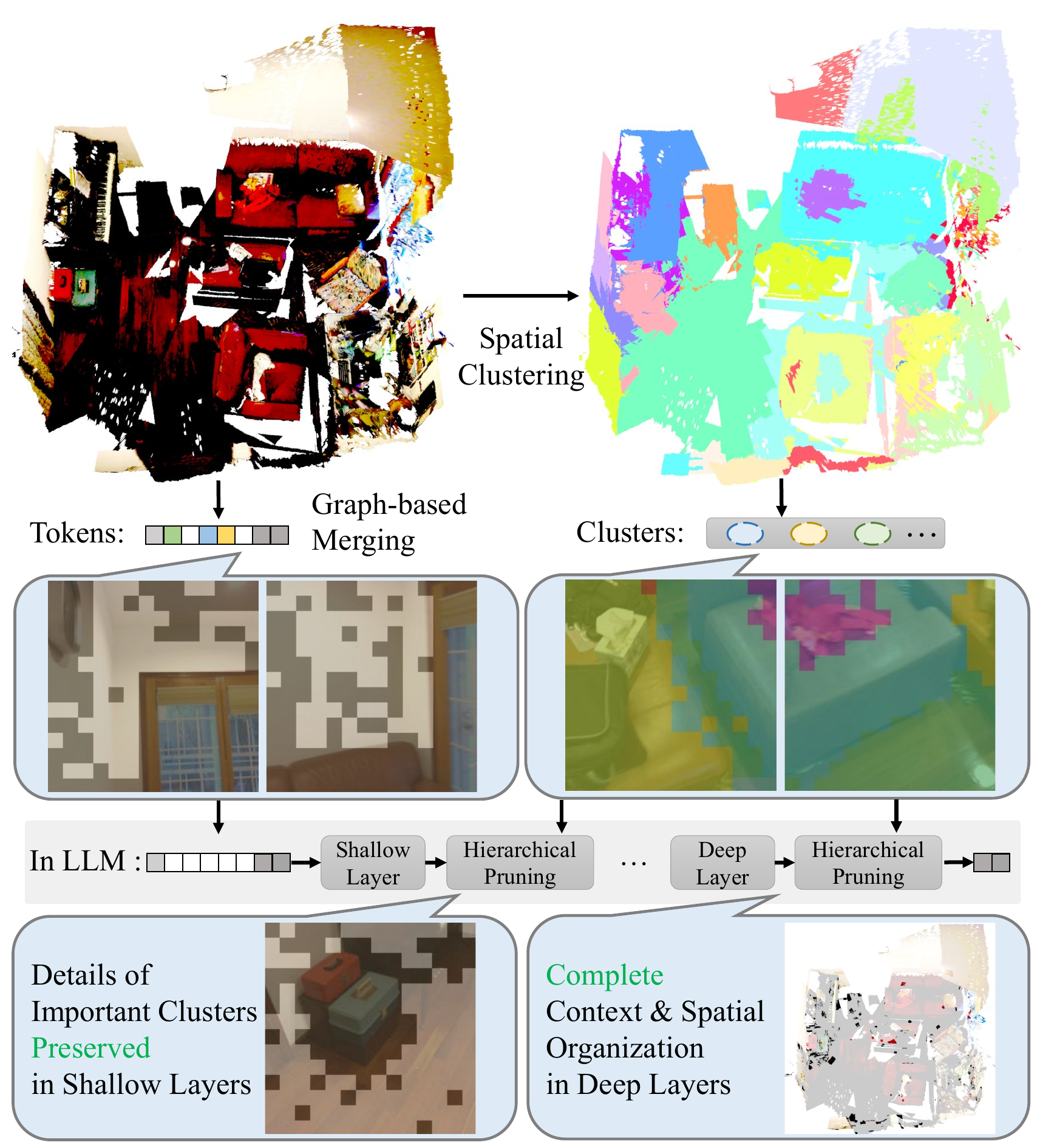}
\caption{Motivation and effects of HiSC. We address structured redundancy in multi-view 3D inputs via complementary designs: graph-based merging spatially connected redundant tokens before LLM, while spatial-semantic clustering enables object-centric hierarchical pruning. Together, they preserve fine-grained details of important objects while maintaining efficiently complete object-level representations.}
\Description{Diagram of HiSC's two-stage token compression.
Graph-based merging consolidates redundant tokens before the large
language model, while spatial clusters guide hierarchical pruning
to retain important-cluster details in shallow layers and scene
context and spatial organization in deep layers.}
\label{fig:intro}
\end{figure}

\section{Introduction}
3D vision-language models (3D VLMs)~\cite{video3dllm,3dllm,llava3d,scenellm} enable unified vision-language-3D reasoning and show strong potential in 3D scene understanding~\cite{ll3da,video3dllm,chatscene} and embodied robotics~\cite{embodiedgpt,occworld,wang2026actionreasoning}. Recent approaches adopt multi-view formulations~\cite{video3dllm,llava3d,jm3d,loopgaussian} that lift 2D image tokens into 3D space, allowing pretrained 2D visual representations to support spatial reasoning.

However, multi-view 3D VLMs incur substantial computational overhead because a single scene often produces thousands of visual tokens~\cite{video3dllm,3dllm}, far exceeding typical 2D VLMs~\cite{llava1p5,qwen2vl}. This inefficiency primarily stems from structured redundancy: uninformative regions contribute excessive tokens, while overlapping views repeatedly encode the same physical entities. The resulting long sequences substantially increase inference cost, posing a key challenge for efficient 3D VLM deployment.

To alleviate the high computational cost, a growing body of work explores training-free token pruning strategies for vision-language models. Early approaches primarily rely on attention-based heuristics, discarding tokens with low cross-modal attention scores~\cite{fastv,fastervlm}. Subsequent methods introduce more principled criteria, such as approximating global attention distributions~\cite{sgl}, estimating token importance via zeroth-order sensitivity~\cite{zooprune}, or enforcing feature diversity during selection~\cite{divprune,btp}. These methods perform token selection directly on the 1D token sequence or feature space and have demonstrated promising efficiency gains in 2D settings. However, such token-level pruning strategies are not well aligned with the requirements of multi-view 3D VLMs. Since tokens are selected independently, existing methods lack explicit mechanisms to ensure that all spatial regions or object instances are adequately represented, potentially leading to missing objects in the compressed input. Meanwhile, importance-based selection tends to concentrate tokens on a small subset of highly relevant regions, which may discard useful contextual details or reduce the representation fidelity of important objects. As a result, it remains challenging to simultaneously preserve scene-level coverage and maintain sufficient detail for reasoning-critical regions under aggressive compression.

In addition, redundancy in multi-view 3D inputs exhibits structured patterns that are not captured by existing pruning paradigms. Unlike 2D images where redundancy mainly arises from local similarity, 3D scenes contain large groups of tokens corresponding to physically consistent regions (e.g., blank planar surfaces) as well as duplicated observations across views. Prior works in 3D vision have shown that such redundancy is closely related to spatial connectivity and geometric consistency~\cite{gitmerge3d,fastvggt}. However, current token pruning methods do not explicitly model these structures, leading to inefficient compression where redundant tokens are retained while informative regions are underrepresented.
These limitations suggest that effective acceleration of 3D VLMs requires moving beyond independent token selection and instead organizing tokens into spatial groups, enabling compression to be performed at different levels with tailored strategies. Such a design allows redundant tokens to be removed in a structured manner while preserving both the coverage of object instances and the detailed information required for downstream reasoning.

To address the above challenges, we present \textbf{HiSC}, a acceleration framework for multi-view 3D VLMs based on \emph{Hierarchical Spatial Clustering Token Compression}, as shown
in Figure~\ref{fig:intro}. The core idea is to organize tokens into spatially grounded clusters using joint geometric proximity and semantic similarity, and perform compression at multiple levels with tailored strategies. Specifically, we first introduce a spatial graph-based token merging strategy that models redundancy as spatial connectivity in 3D space. By constructing a connectivity graph over tokens and grouping physically consistent regions, this design enables structured consolidation of large planar surfaces and cross-view duplicated observations before LLM inference. Such cluster-level redundancy removal significantly reduces token count while preserving the completeness of spatial regions, providing a compact yet faithful representation of the scene.

While removing structured redundancy is essential, effectively preserving informative content during aggressive compression remains a key challenge. We therefore propose a spatial clustering-based pruning paradigm that performs hierarchical compression within LLM inference. Instead of selecting tokens independently, our method treats clusters as the basic units and applies level-specific strategies across and within clusters. At the cluster level, tokens are allocated to ensure object-level coverage across the scene, while at the intra-cluster level, more tokens are retained for regions that are more relevant to the reasoning process. This design enables the model to maintain both the completeness of object-level spatial information and the fidelity of fine-grained details, resulting in a better trade-off between efficiency and reasoning performance.

Extensive experiments across diverse 3D reasoning tasks demonstrate that HiSC achieves state-of-the-art performance while significantly improving efficiency. Notably, our method maintains over 90\% of the original performance under extreme compression ratios, and remains nearly lossless under moderate compression settings. In summary, our main contributions are as follows:

\begin{itemize}
[leftmargin=0.4cm, itemindent=0cm]
\item We propose \textbf{HiSC}, a training-free framework for token compression in 3D VLMs, which lifts token compression from token-level selection to cluster-level processing by organizing tokens into spatial clusters using joint geometric and semantic cues.

\item We introduce a \textbf{spatial graph-based merging (SGraM) strategy} that models cross-view redundancy as spatial connectivity and aggregates physically consistent regions, effectively merging redundant tokens prior to LLM inference.

\item We propose a \textbf{spatial clustering-based pruning (SCluP) paradigm} within LLM, which performs compression across and within clusters, preserving object-level completeness while retaining fine-grained details in important regions.
\end{itemize}

\section{Related Works}

\subsection{3D Vision-Language Models.}
With advances in deep learning~\cite{zhang2026real, liang2025relation, liang2025Flow, zhang2025unaligned, ZhangTao2024CJE, zhang2026enhancing, zhang2024deep, zhang2026supervise} and Transformers~\cite{li2023joint, TianYe2023CJE, liang2026Lift3Dreamer}, vision-language models have rapidly evolved~\cite{qwen2vl,llavavideo}.
Recent years have witnessed a paradigm shift from explicit 3D representations (e.g., point clouds) to multi-view image-based 3D vision-language models (3D-LLMs). Early efforts such as 3D-LLM~\cite{3dllm} and PointLLM~\cite{pointllm} directly operate on point clouds with dedicated 3D encoders, enabling dense captioning and spatial reasoning but suffering from high computational cost and limited compatibility with large-scale 2D pretraining. To bridge this gap, recent works adopt multi-view images augmented with 3D positional cues~\cite{seqvlm}. For instance, Video-3D LLM~\cite{video3dllm} models 3D scenes as video sequences with injected 3D position encoding, while LLaVA-3D~\cite{llava3d} constructs 3D-aware tokens by lifting 2D patches into 3D space via camera geometry and depth signals. Such designs effectively inherit strong 2D priors from pretrained vision encoders (e.g., CLIP~\cite{clip}) while enabling spatial reasoning. However, the multi-view formulation introduces extremely long visual token sequences, making inference expensive. Existing solutions (e.g., sampling or pooling) often rely on heuristic compression~\cite{tome,btp}, which may discard fine-grained spatial details critical for downstream 3D question answering tasks.

\subsection{Acceleration for Vision-Language Models.}
While alternative directions such as model quantization and adaptive compression have been explored~\cite{mquant,fast3d}, a growing body of work focuses on training-free acceleration for VLMs, particularly through token or structure pruning. Early methods exploit attention sparsity~\cite{evovit,dynamicvit,lvlmcsp}, where FastV~\cite{fastv} and FasterVLM~\cite{fastervlm} remove tokens based on attention signals at different stages, and SGL~\cite{sgl} approximates global attention to guide pruning. Beyond token-level selection, Short-LVLM~\cite{shortlvlm} explores layer-wise redundancy by pruning unimportant layers, while SparseVLM~\cite{sparsevlm} identifies and retains only the most informative regions through dynamic sparsification, striking a balance between detail and efficiency.

\begin{figure*}[!t]
\centering
\includegraphics[width=0.99\linewidth]{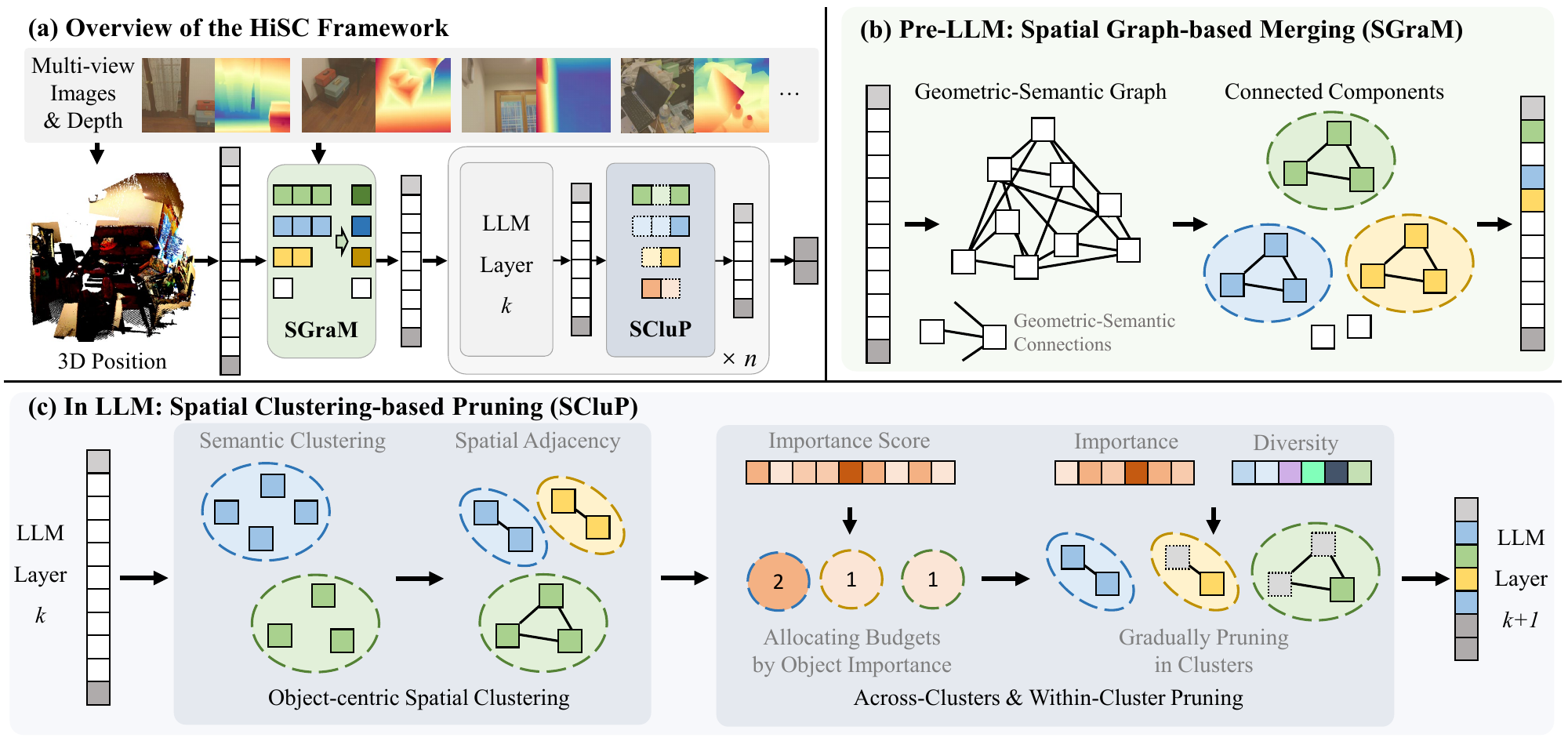}
\caption{Overview of the \textbf{HiSC} framework. (a) Overall pipeline showing graph-based token merging before LLM inference and hierarchical spatial clustering-based pruning in LLM. (b) \textbf{Spatial Graph-based Merging} groups and merges spatially connected redundant tokens to preserve consistent regions. (c) \textbf{Hierarchical Spatial Clustering-based Pruning} organizes tokens into object-centric clusters and performs structured allocation, ensuring object-level coverage while retaining fine-grained details.}
\Description{Three-panel architecture of HiSC. Multi-view images,
depth, and three-dimensional positions form visual tokens;
spatial graph-based merging aggregates geometric-semantic connected
components before the large language model, while spatial
clustering-based pruning allocates cluster budgets by importance
and selects important and diverse tokens between layers.}
\label{fig:main}
\end{figure*}

More recent works revisit pruning from a more principled perspective, incorporating criteria such as importance, diversity, and information flow~\cite{tokenlearner,efficientcontent}. ZOO-Prune~\cite{zooprune}, VisionZip~\cite{visionzip} and DivPrune~\cite{divprune} emphasize token sensitivity and diversity, while Balanced-Token-Pruning~\cite{btp} and VFlowOpt~\cite{vflowopt} model layer-wise trade-offs and information preservation during pruning. Other approaches such as VisPruner~\cite{vispruner} and VisionTrim~\cite{visiontrim} further improve robustness via encoder-side filtering or coordinated multi-stage compression. Despite their effectiveness, these methods are primarily designed for 2D inputs and overlook geometric consistency across views, limiting their applicability to multi-view 3D settings.

Beyond VLMs, the 3D vision community has explored efficient Transformer designs for large-scale spatial data. Point Transformer V3~\cite{ptv3} replaces costly KNN operations with serialized neighbor mapping, significantly improving scalability. GitMerge3D~\cite{gitmerge3d} reveals severe token redundancy in 3D Transformers and proposes a globally-informed graph merging strategy. For multi-view geometry tasks, FastVGGT~\cite{fastvggt} identifies the token collapse phenomenon caused by redundant view overlap and introduces a training-free merging pipeline with anchor token preservation, salient token bypass, and region-based sampling. These methods highlight that geometric redundancy differs fundamentally from semantic redundancy, and preserving spatial anchors is crucial for 3D reasoning.

\section{Method}
\label{sec:method}

We first introduce the formulation of multi-view 3D VLMs and describe how visual tokens are constructed and consumed during inference in Section~\ref{sec:prelim}. Based on this formulation, we present \textbf{HiSC}, our framework for hierarchical spatial clustering token compression in 3D VLMs, as shown in Figure~\ref{fig:main}. Specifically, Section~\ref{sec:merge} introduces a graph-based token merging strategy that removes structured redundancy prior to LLM inference, and Section~\ref{sec:prune} presents a hierarchical spatial clustering-based pruning scheme within LLM inference.

\subsection{Preliminary}
\label{sec:prelim}

We consider a representative class of multi-view 3D VLMs that take as input multi-view RGB images with corresponding depth maps and camera parameters, and perform 3D scene reasoning via spatially grounded visual tokens.

\subsubsection{\textbf{Visual Token Construction.}}
Given multi-view RGB images with corresponding depth maps and camera parameters, each image is encoded into patch-level visual features. Using depth and camera geometry, each patch is associated with a 3D coordinate. Spatially grounded visual tokens $\mathbf{v}_i$ are constructed by injecting spatial information into visual features as 
$
\mathbf{v}_i = f_{\text{proj}}(\mathbf{f}_i, \mathbf{p}_i) \in \mathbb{R}^{d},
$
where $\mathbf{f}_i \in \mathbb{R}^{d_f}$ denotes the visual feature and $\mathbf{p}_i \in \mathbb{R}^{3}$ denotes the corresponding 3D coordinate of the $i$-th token. The projection function $f_{\text{proj}}(\cdot)$ integrates appearance and spatial information, e.g., by applying a positional encoding $\phi(\mathbf{p}_i)$ and combining it with $\mathbf{f}_i$.

\subsubsection{\textbf{Multi-modal Input.}}
All visual tokens are stacked into a token matrix 
$\mathbf{V} \in \mathbb{R}^{L_v \times d}$,
which is combined with textual tokens, including system prompt $\mathbf{T}_{\text{prompt}}$ and query tokens $\mathbf{T}_{\text{query}}$, to form the input to the LLM:
\begin{equation}
\mathbf{X} = [\mathbf{T}_{\text{prompt}}; \mathbf{V}; \mathbf{T}_{\text{query}}] \in \mathbb{R}^{(L_v + L_{\text{text}}) \times d}.
\end{equation}
where the number of visual tokens $L_v$ is typically much larger than that of textual tokens $L_{text}$, and dominates the computational cost, due to the multi-view settings. Moreover, these tokens contain redundancy due to overlapping observations and spatially consistent regions, making efficient token compression crucial for 3D VLM.

\subsection{Spatial Graph-based Merging}
\label{sec:merge}

Multi-view 3D inputs contain substantial structured redundancy due to repeated observations and large physically consistent regions, where many tokens are near-duplicate and carry almost identical information, making pre-inference merging desirable to reduce computation and attention dilution. A natural approach is to merge tokens based on feature similarity, with or without 3D positional encoding; however, such methods rely on global similarity and fail to respect 3D structure: tokens from different objects may be incorrectly merged due to appearance similarity, while tokens from the same surface may be separated under spatial variation. More fundamentally, they lack an explicit notion of \emph{connectivity}—physically consistent regions are spatially connected through chains of locally adjacent and semantically consistent tokens, a transitive property that similarity-based clustering cannot capture, leading to either over-segmentation or erroneous merging.

To address this issue, we propose a graph-based geometric-semantic redundancy merging strategy that models spatial connectivity while preserving semantic consistency. Specifically, we decouple geometric proximity and semantic similarity to construct a connectivity graph, where edges are established only between tokens that are spatially adjacent and highly similar in appearance. This formulation naturally captures the transitive connectivity of physically consistent regions, enabling tokens with large spatial variation but consistent local structure to be grouped, while avoiding incorrect aggregation across disconnected regions.

\subsubsection{\textbf{Geometric-Semantic Adjacency.}}
To identify reliable local relations, we explicitly decouple geometric proximity and semantic similarity when constructing token connections. Given visual tokens $\{\mathbf{v}_i\}_{i=1}^{L_v}$ with associated 3D positions $\{\mathbf{p}_i\}_{i=1}^{L_v}$, we define geometric adjacency $\mathbf{A}^{\text{geo}}_{ij}$ and semantic adjacency $\mathbf{A}^{\text{sem}}_{ij}$ as:
\begin{equation}
\mathbf{A}^{\text{geo}}_{ij} = \mathbb{I}(\|\mathbf{p}_i - \mathbf{p}_j\|_2 < \tau_{\text{geo}}), \quad
\mathbf{A}^{\text{sem}}_{ij} = \mathbb{I}(\text{sim}(\mathbf{f}_i, \mathbf{f}_j) > \tau_{\text{sem}}),
\end{equation}
where $\|\cdot\|_2$ is the Euclidean distance; $\text{sim}(\cdot,\cdot)$ denotes cosine similarity; $\tau_{\text{geo}}$ and $\tau_{\text{sem}}$ are thresholds for geometric proximity and semantic similarity; and $\mathbb{I}(\cdot)$ is the indicator function that outputs 1 if the condition holds and 0 otherwise. Then we combine them into a unified adjacency as $\mathbf{A}_{ij} = \mathbf{A}^{\text{geo}}_{ij} \odot \mathbf{A}^{\text{sem}}_{ij}$. This ensures that edges are established only between tokens that are both spatially adjacent and semantically consistent, filtering out spurious matches caused by either appearance ambiguity or spatial overlap.

This decoupled design provides special constraints enforce spatial validity, while semantic constraints prevent merging across distinct structures. This lays the foundation for constructing meaningful connectivity patterns in 3D scenes.

\subsubsection{\textbf{Connectivity-based Connected Groups.}}
To capture the transitive connectivity of physically consistent regions, we interpret the adjacency matrix $\mathbf{A}$ as an undirected graph 
$\mathcal{G} = (\mathcal{V}, \mathcal{E})$ and extract its connected components:
\begin{equation}
\mathcal{C} = \{\mathcal{C}_k\}_{k=1}^{K}, \quad 
\mathcal{C}_k \subseteq \mathcal{V}, \quad 
\mathcal{C}_k \cap \mathcal{C}_{k'} = \emptyset
\end{equation}
where $\mathcal{V}$ and $\mathcal{E}$ denote the sets of nodes (tokens) and edges defined by the adjacency matrix $\mathbf{A}$, respectively; $\mathcal{C}$ represents the set of connected components, $K$ is the total number of components, and each $\mathcal{C}_k$ denotes a subset of tokens forming a connected component. Each component corresponds to a \emph{connected domain} that approximates a physically consistent region in the scene.

Finally, tokens within each connected component are consolidated into a representative super-token via simple aggregation, such as feature and position averaging, which removes redundancy while preserving the overall spatial layout and semantic consistency. The resulting set of super-tokens forms a compact yet faithful representation of the scene for subsequent LLM inference.

\subsection{Spatial Clustering-based Pruning}
\label{sec:prune}

Although redundancy can be reduced before LLM inference, visual tokens still exhibit significant redundancy during multi-layer reasoning, making layer-wise pruning necessary. However, existing token-level methods treat tokens independently and ignore the structured nature of 3D scenes, often leading to imbalanced token allocation, including over-pruning spatially coherent regions while failing to preserve fine-grained details for important objects.

To address this, we propose a \emph{hierarchical spatial clustering-based token pruning} paradigm, which performs compression over structured spatial units instead of individual tokens. Specifically, tokens are first grouped into object-centric clusters, followed by progressive layer-wise selection via cluster-level allocation and intra-cluster sampling. This enables better coordination between global coverage and local detail preservation.

\subsubsection{\textbf{Object-centric Spatial Clustering.}}
To enable structured token pruning aligned with the physical organization of 3D scenes, we first group tokens into object-centric clusters that serve as basic units for hierarchical compression. Given visual tokens $\{\mathbf{v}_i\}_{i=1}^{L_v}$ with positions $\{\mathbf{p}_i\}_{i=1}^{L_v}$, we decompose clustering into two steps: semantic grouping followed by spatial instance separation. Specifically, tokens are first partitioned into coarse semantic sets, and then spatial connectivity is enforced within each set to separate physically distinct instances. The final clusters are obtained as:
\begin{equation}
\mathcal{C} = \bigcup_{m=1}^{K_{\text{sem}}} \mathrm{ConnComp}(\mathbf{A}^{(m)}),\quad
\mathbf{A}^{(m)} \in \{0,1\}^{|\mathcal{S}_m| \times |\mathcal{S}_m|}
\end{equation}
where $\mathcal{C} = \{\mathcal{C}_k\}$ denotes the set of clusters, $K_{\text{sem}}$ is the number of semantic groups, $\mathbf{A}^{(m)}$ is the spatial adjacency defined within the $m$-th semantic group, $\mathcal{S}_m$ denotes the set of tokens belonging to the $m$-th semantic group, and $\mathrm{ConnComp}(\cdot)$ extracts the connected components of a graph. Each cluster $\mathcal{C}_k$ thus corresponds to a physically connected region in 3D space.

This design decouples semantic similarity from spatial continuity to model object-level structures in 3D scenes. Purely semantic grouping tends to merge visually similar but spatially distant regions, while purely spatial partitioning fails to capture semantic identity. By first enforcing semantic consistency and then applying spatial connectivity, our formulation ensures that each cluster corresponds to a physically coherent region. This avoids incorrect merging across disconnected objects while preventing over-segmentation within the same object. Moreover, the connectivity-based decomposition preserves extended surfaces and cross-view overlaps, producing clusters that align well with objects and provide a reliable basis for subsequent structured token allocation.

\subsubsection{\textbf{Hierarchical Cluster-based Token Pruning.}}
Based on the obtained clusters $\mathcal{C} = \{\mathcal{C}_k\}$, we perform structured token pruning by allocating token budgets across clusters and selecting tokens within each cluster. Let $N^{(l)}$ denote the number of tokens retained at layer $l$. Instead of selecting tokens independently, we allocate budgets $\{n_k\}$ at the cluster level:
$
n_k \geq 1, \quad n_k \propto (I_k)^2, \quad \sum_k n_k = N^{(l)},
$
where $n_k$ is the number of tokens assigned to cluster $\mathcal{C}_k$, and $I_k$ denotes the importance score of cluster $\mathcal{C}_k$ computed from its tokens. The constraint $n_k \geq 1$ ensures that each cluster is preserved, maintaining spatial coverage and preventing object disappearance, while the squared weighting emphasizes important regions.

Within each cluster, tokens are further selected to satisfy the allocated budget. We adopt a hybrid strategy that combines importance-based selection with diversity-aware sampling. To encourage structural diversity, we define a joint spatial-semantic distance:
\begin{equation}
D_{ij}^{\text{comb}} = \lambda \|\mathbf{p}_i - \mathbf{p}_j\|_2 
+ (1-\lambda)\big(1 - \text{sim}(\mathbf{f}_i, \mathbf{f}_j)\big),
\end{equation}
where $\mathbf{p}_i$ and $\mathbf{f}_i$ denote the position and feature of token $i$, $\|\cdot\|_2$ is the Euclidean distance, $\text{sim}(\cdot, \cdot)$ is cosine similarity, and $\lambda \in [0,1]$ balances geometric and semantic terms. Tokens that are distant from already selected ones under this metric are preferred, promoting coverage of diverse structures.

Overall, this hierarchical pruning scheme aligns token compression with the structured nature of 3D scenes. By performing allocation at the cluster level and applying progressive importance-diversity pruning within each cluster, the method explicitly ensures object-level coverage while preserving fine-grained details of important regions for deeper reasoning. This design avoids common failure modes of token-level pruning: early-stage pruning may discard detailed information of important objects due to diversity-oriented selection, while later-stage aggressive pruning on less important regions can lead to incomplete object-level representation. By coordinating global allocation and intra-cluster refinement, our method maintains both spatial completeness and detailed representation, enabling robust performance under high compression.

\begin{table*}[t!]
    \centering
    \caption{Performance comparison on Video-3D-LLM~\cite{video3dllm} under different token retention conditions. Avg. Tokens denotes the average number of vision tokens across LLM layers, and TFLOPS denotes the computational cost during LLM inference. Relative Score refers to the average percentage of performance retained across 5 benchmarks compared to the original LLM. }
    \begin{threeparttable}
    \renewcommand{\arraystretch}{1.03} 
    \setlength{\tabcolsep}{3pt}
\resizebox{\textwidth}{!}{

    \begin{tabular}{l|cc|ccccccccc|c}
        \toprule 
        \multirow{2}{*}{Method} & \multirow{2}{*}{\makecell{Avg. \\Tokens$\downarrow$}} & \multirow{2}{*}{\makecell{TFLOPS$\downarrow$}} & \multicolumn{2}{c}{ScanRefer} & \multicolumn{2}{c}{Multi3DRef} & \multicolumn{2}{c}{Scan2Cap}  & \multicolumn{2}{c}{ScanQA} & \multicolumn{1}{c}{SQA3D} &  \multirow{2}{*}{\makecell{Relative \\Score}}  \\
        \cmidrule(lr){4-5} \cmidrule(lr){6-7} \cmidrule(lr){8-9} \cmidrule(lr){10-11} \cmidrule(lr){12-12}
        & & & \scalebox{0.85}[1]{Acc@0.25} & \scalebox{0.85}[1]{Acc@0.5} & \scalebox{0.9}[1]{F1@0.25} & \scalebox{0.9}[1]{F1@0.5} & \scalebox{0.85}[1]{B-4@0.5} & \scalebox{0.85}[1]{C@0.5} & C & EM & EM & \\ \midrule
        \rowcolor{lightgray} \multicolumn{13}{c}{\textit{Upper Bound, 100\% Tokens}} \\
        Vanilla & 100.0\% & 95.66 & 58.20 & 51.79 & 57.40 & 52.13 & 41.19 & 83.97 & 102.11 & 30.10 & 58.45 & 100\% \\
        \midrule
        \rowcolor{lightgray} \multicolumn{13}{c}{\textit{Light Compression (High Budget, $\sim$50\% Tokens)}} \\
        FastV~\cite{fastv}\texttt{\scriptsize{(ECCV24)}} & 53.6\% & 49.78 & 57.16 & \underline{51.17} & 55.75 & 50.41 & 36.97 & 67.84 & 96.26 & 28.34 & 57.97 & 94.33\%\\
        VFlowOpt~\cite{vflowopt}\texttt{\scriptsize{(ICCV25)}} & 51.4\% & 51.29 & 57.35	 & 51.09 & \textbf{56.91} & \textbf{51.76} & \underline{40.35} & \underline{79.01} & 100.10 & 29.58 & \underline{58.06} & \underline{98.15\%}\\
        VisPruner~\cite{vispruner}\texttt{\scriptsize{(ICCV25)}} & \underline{50.0\%} & \underline{48.16} & 
        57.28 & 50.94 & 56.19 & 50.67 & 39.90 & 76.82 & 98.88 & 29.22 & 57.94 & 97.03\%\\
        VisionTrim~\cite{visiontrim}\texttt{\scriptsize{(ICLR26)}} & 54.9\% & 59.90 & 
        56.73 & 50.31 & 56.17 & 50.97 & 39.18 & 75.90 & 96.55 & 28.90 & 57.80 & 96.13\%\\
        FastVGGT~\cite{fastvggt}\texttt{\scriptsize{(ICLR26)}}\tnote{1} & 53.6\% & 52.74 & 
        \underline{57.47} & 51.09 & 56.61 & 51.38 & 39.11 & 74.23 & \underline{100.31} & \underline{29.90} & 57.46 & 97.09\%\\
        \textbf{Ours} & \textbf{49.2\%} & \textbf{47.93} & \textbf{57.90} & \textbf{51.47} & \underline{56.82} & \underline{51.67} & \textbf{40.53} & \textbf{80.57} & \textbf{101.21} & \textbf{29.95} & \textbf{58.37} & \textbf{98.87\%}\\
        \midrule
        \rowcolor{lightgray} \multicolumn{13}{c}{\textit{Medium Compression (Mid Budget, $\sim$25\%  Tokens)}} \\
        FastV~\cite{fastv}\texttt{\scriptsize{(ECCV24)}} & 26.8\% & 26.80 & 56.53 & \underline{50.56} & 46.74 & 42.95 & 35.64 & 59.14 & 92.61 & 26.35 & 51.69 & 86.91\%\\
        VFlowOpt~\cite{vflowopt}\texttt{\scriptsize{(ICCV25)}} & 26.1\% & 29.15 & 56.75 & 50.50 & \underline{55.37} & \underline{50.43} & \underline{39.40} & \underline{73.40} & 96.30 & 28.45 & \underline{57.49} & \underline{95.39\%}\\
        VisPruner~\cite{vispruner}\texttt{\scriptsize{(ICCV25)}} & \underline{25.0\%} & \underline{24.46} & 
        56.70 & 49.97 & 54.81 & 49.05 & 38.10 & 70.68 & \underline{96.63} & \underline{28.66} & 57.09 & 94.19\%\\
        BTP~\cite{btp}\texttt{\scriptsize{(NeurIPS25)}}\tnote{2} & 27.2\% & 25.91 & 
        \underline{56.83} & 50.48 & 55.28 & 50.15 & 36.74 & 64.71 & 85.81 & 23.79 & 54.16 & 89.96\%\\
        VisionTrim~\cite{visiontrim}\texttt{\scriptsize{(ICLR26)}} & 34.6\% & 37.71 & 
        55.90 & 49.59 & 55.25 & 50.21 & 37.30 & 65.96 & 92.46 & 27.49 & 56.86 & 92.51\%\\
        \textbf{Ours} & \textbf{24.7\%} & \textbf{24.06} & \textbf{57.20} & \textbf{50.89} & \textbf{55.75} & \textbf{50.77} & \textbf{39.56} & \textbf{76.01} & \textbf{100.78} & \textbf{29.82} & \textbf{57.77} & \textbf{97.14\%}\\
        \midrule
        \rowcolor{lightgray} \multicolumn{13}{c}{\textit{Extreme Compression (Low Budget, $\sim$10\%  Tokens)}} \\
        FastV~\cite{fastv}\texttt{\scriptsize{(ECCV24)}} & 10.7\% & 15.84 & 54.64 & 48.03 & 36.90 & 34.19 & 32.59 & 45.16 & 83.40 & 23.47 & 50.24 & 77.22\%\\
        VFlowOpt~\cite{vflowopt}\texttt{\scriptsize{(ICCV25)}} & 10.3\% & 15.34 & \underline{55.52} & \underline{49.30} & 50.64 & 46.24 & 35.95 & 60.19 & 90.46 & \underline{26.55} & 54.96 & 88.59\%\\
        VisPruner~\cite{vispruner}\texttt{\scriptsize{(ICCV25)}} & \underline{10.0\%} & \underline{10.18} & 
        55.24 & 48.73 & \underline{52.06} & \underline{46.53} & \underline{36.73} & \underline{62.44} & \underline{90.67} & 26.50 & \underline{55.84} & \underline{89.43\%}\\
        VisionTrim~\cite{visiontrim}\texttt{\scriptsize{(ICLR26)}} & 19.8\% & 24.38 & 
        55.15 & 48.81 & 51.63 & 46.28 & 34.24 & 51.60 & 87.27 & 25.75 & 54.53 & 86.29\%\\
        \textbf{Ours} & \textbf{9.8\%} & \textbf{9.90} & \textbf{56.01} & \textbf{49.82} & \textbf{52.40} & \textbf{47.71} & \textbf{37.59} & \textbf{67.95} & \textbf{96.21} & \textbf{28.36} & \textbf{56.27} & \textbf{92.46\%}\\
        \bottomrule
    \end{tabular}
    }
    \begin{tablenotes} 
        \footnotesize    
        \item[1] FastVGGT is unable to achieve higher-rate compression acceleration due to the limitations of its method.
        \item[2] BTP employs a fixed ratio for token compression acceleration. We only compare its original settings. 
      \end{tablenotes}  
    \end{threeparttable}    
    \label{tab:main_results}
\end{table*}

\section{Experiment}

\noindent In this section, we comprehensively evaluate the effectiveness of our proposed method. We first introduce the experimental setup, including datasets, metrics, and implementation details. We then report quantitative results in comparison with prior methods, along with qualitative visualizations for better interpretation. Finally, we conduct in-depth analyses and ablation studies to examine the design choices and individual contributions of our framework.

\subsection{Experimental Setup}

\subsubsection{\textbf{Tasks and Datasets.}}
We evaluate our method on a diverse set of 3D scene understanding tasks built upon multi-view RGB-D observations, following the standard setup of Video-3D LLMs. Specifically, we consider visual grounding, dense captioning, and 3D question answering, which collectively require both accurate spatial localization and holistic scene understanding under multi-view inputs. We adopt five widely used benchmarks from the ScanNet~\cite{scannet} dataset, including ScanRefer~\cite{scanrefer} and Multi3DRefer~\cite{multi3drefer} for object grounding, Scan2Cap~\cite{scan2cap} for dense captioning, and ScanQA~\cite{scanqa} and SQA3D~\cite{sqa3d} for question answering. These datasets provide richly annotated indoor scenes with aligned RGB-D video frames and camera parameters, making them well-suited for evaluating multi-view 3D reasoning. Following prior work~\cite{video3dllm}, we use the standard validation splits for ScanRefer, Multi3DRefer, Scan2Cap, and ScanQA, and the test split for SQA3D.

\subsubsection{\textbf{Evaluation Metrics.}}
We adopt standard metrics to comprehensively evaluate both spatial accuracy and semantic quality. For grounding tasks, we report Acc@0.25 and Acc@0.5 on ScanRefer~\cite{scanrefer}, and F1 scores at IoU thresholds of 0.25 and 0.5 for Multi3DRefer~\cite{multi3drefer}, measuring the correctness of predicted object localization. For dense captioning on Scan2Cap~\cite{scan2cap}, we use CIDEr@0.5IoU and BLEU-4@0.5IoU, which jointly assess linguistic quality and spatial alignment. For question answering, we report CIDEr and exact match (EM) on ScanQA~\cite{scanqa}, and EM on SQA3D~\cite{sqa3d}. These metrics not only reflect the reasoning capability, but also serve as indicators of how well compressed tokens preserve spatial coverage and fine-grained details under aggressive token reduction.

\subsubsection{\textbf{Implementation Details}}
We apply our HiSC framework to a representative multi-view 3D VLM, built upon the LLaVA-Video 7B~\cite{llavavideo} architecture. We directly use the publicly released pretrained weights and follow its default configuration for visual encoding and multi-modal input construction. Our acceleration modules are inserted at inference time without any retraining or fine-tuning. Unless otherwise specified, we adopt $K_{\text{sem}}=16$ semantic clusters for object-centric grouping, followed by spatial connected component splitting with threshold $\tau_{\text{dist}}$ set to $1/28$ of the scene span, as images are divided into 14 by 14 patches. For hierarchical pruning, we adopt the selection strategy of the BTP~\cite{btp} method, and perform layer-wise token selection at predefined layers $\{1,4,7,16,23\}$, with a target average compression ratio $P_{\text{avg}}$ enforced through a progressive retention schedule controlled by decay factor $\beta=0.01$. Token importance is computed via a depth-aware fusion of cross-modal similarity and LLM attention, and cluster-level budgets are allocated proportionally to squared importance with a minimum retention of one token per cluster. All experiments are conducted on 4 NVIDIA RTX 3090 GPUs with 24GB VRAM, with total inference time of approximately 8h for the original 3D VLM.

\begin{figure*}[!t]
\centering
\includegraphics[width=1.0\linewidth]{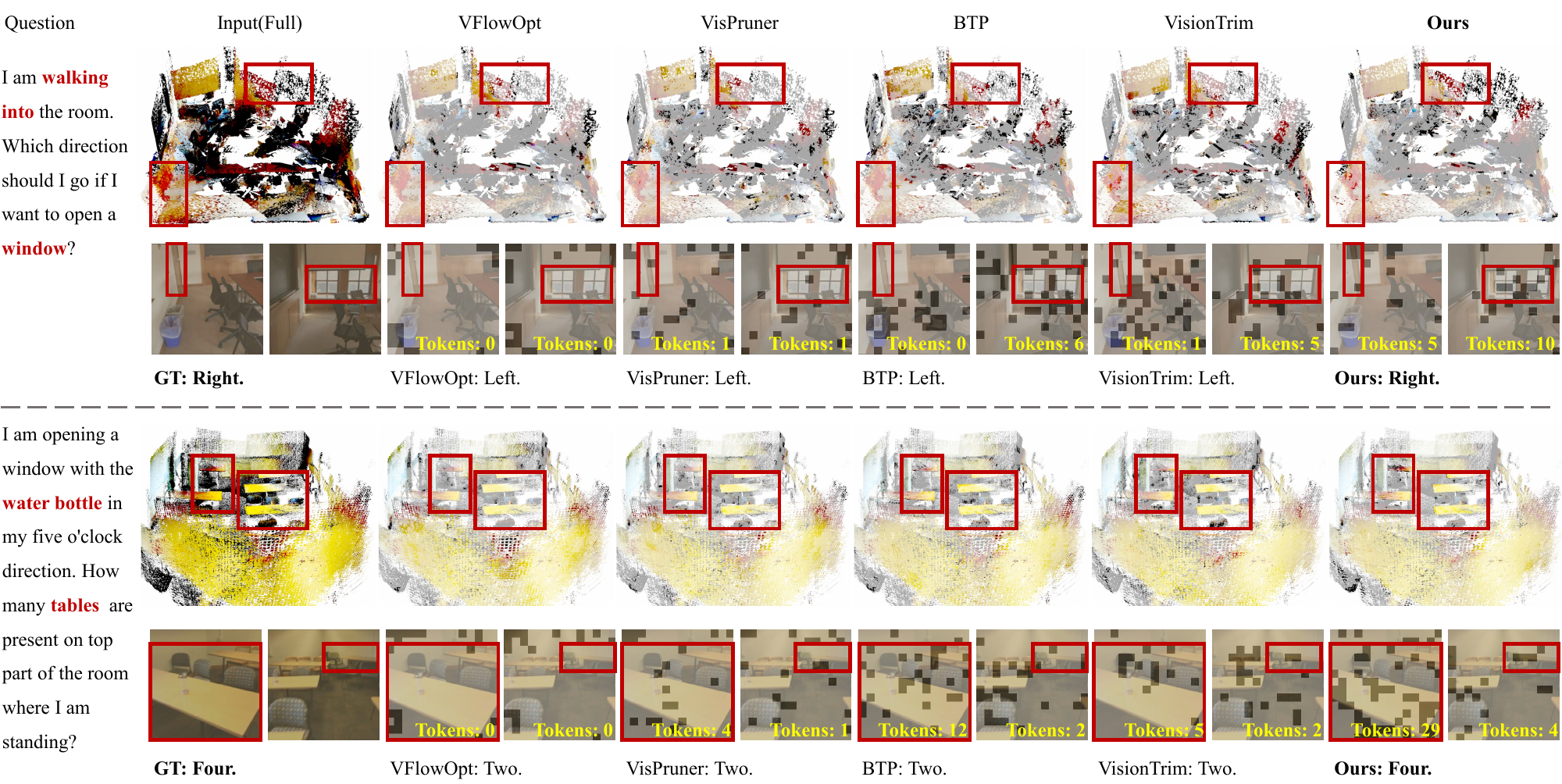}
\caption{Qualitative visualization of retained token distributions under 90\% compression. We compare HiSC with representative methods in 3D scenes (upper) and key 2D views (lower). \textcolor{red}{Red} boxes indicate task-relevant regions, and highlighted areas (with yellow counts) denote retained tokens. Compared to prior methods, HiSC preserves more details in important regions while compactly representing redundant areas, and distinguishes multiple similar objects.}
\Description{Comparison of retained-token distributions for the full
input, representative baselines, and HiSC. Unlike the concentrated or fragmented coverage of prior methods, HiSC preserves broader object-level coverage and more details in task-relevant regions, producing correct predictions.}
\label{fig_vis_main}
\end{figure*}

\subsection{Experimental Results}

\subsubsection{\textbf{Comparing with the-state-of-the-arts.}}
We evaluate the effectiveness of HiSC under different compression regimes and compare it with representative training-free token pruning methods on multi-view 3D reasoning tasks. All experiments are conducted based on a Video-3D LLM as the language backbone, ensuring a unified evaluation setting. We compare against several prior approaches with different pruning strategies, including early attention-based pruning (FastV~\cite{fastv}), layer-wise balanced pruning (BTP~\cite{btp}), information flow optimization (VFlowOpt~\cite{vflowopt}), attention-guided visual pruning with redundancy removal (VisPruner~\cite{vispruner}), and unified pre-decoding compression (VisionTrim~\cite{visiontrim}). We also include FastVGGT~\cite{fastvggt}, originally designed for efficient multi-view 3D reconstruction, by adapting its token reduction strategy to the 3D VLM setting under standard early-layer pruning protocols. For methods with restricted pruning schemes, we follow their default configurations for fair comparison. We evaluate all methods under Light, Medium, and Extreme compression settings with average token reduction ratios of 50\%, 75\%, and 90\%, respectively. 

As shown in Table~\ref{tab:main_results}, HiSC consistently outperforms all baselines across tasks and compression levels. Notably, under light and medium compression, HiSC retains nearly all performance of the uncompressed model, indicating that removing structured redundancy can alleviate attention dilution. Under extreme compression, our method maintains significantly higher performance than prior approaches, demonstrating its ability to preserve object-level completeness and fine-grained details under aggressive token reduction.

\subsubsection{\textbf{Qualitative Visualization of Token Compression.}}
Figure~\ref{fig_vis_main} presents qualitative visualizations of token distributions before and after compression, comparing HiSC with representative token pruning methods. We visualize the spatial distribution of retained tokens in 3D scenes under 90\% compression ratio to highlight how different methods preserve scene structure. Existing token-level pruning approaches tend to either over-concentrate tokens or produce fragmented coverage, leading to incomplete object representations and loss of spatial coverage. In contrast, HiSC maintains object-level structures: redundant tokens from large homogeneous regions and cross-view overlaps are effectively consolidated, while tokens are distributed in an object-centric manner that preserves both global scene coverage and local details. Notably, important objects retain dense and diverse token representations, while less informative regions are compactly summarized. These results show that our graph-based merging and hierarchical cluster-based pruning jointly enable structured compression, aligning token reduction with the geometric and semantic organization of 3D scenes and preserving critical information even under aggressive compression.

\subsection{Discussions}

\subsubsection{\textbf{Effect of Geometric-Semantic Connectivity Design.}}
We analyze the graph-based merging strategy by comparing semantic-only connectivity, geometric-only connectivity, and connectivity based on entangled 3D-aware features under 90\% compression. As reported in Table~\ref{ablation_merge}, semantic-only and geometric-only variants cause substantial performance drops because they cannot jointly enforce spatial continuity and semantic consistency among redundant tokens. Although 3D-encoded features partially alleviate this issue, they still underperform our decoupled formulation, as entangling spatial and appearance information reduces similarity reliability. In contrast, our geometric-semantic connectivity models local consistency in both domains, enabling more accurate identification and merging of redundant tokens. This confirms the necessity of decoupled connectivity for robust pre-inference merging.

\begin{table}[t!]
    \centering
    \caption{Comparison of different adjacency constructions.}
    \renewcommand{\arraystretch}{1.0}
    \setlength{\tabcolsep}{2pt}
\resizebox{\columnwidth}{!}{
    \begin{tabular}{l|cccccc}
        \toprule 
        \multirow{2}{*}{Setting} & \multicolumn{2}{c}{ScanRefer} & \multicolumn{2}{c}{Scan2Cap}  & \multicolumn{2}{c}{ScanQA} \\ 
         \cmidrule(lr){2-3} \cmidrule(lr){4-5} \cmidrule(lr){6-7}
        & \scalebox{0.9}[1]{Acc@0.25} & \scalebox{0.9}[1]{Acc@0.5} & \scalebox{0.85}[1]{B-4@0.5} & \scalebox{0.85}[1]{C@0.5} & C & EM \\ \midrule
        semantic-only & 54.99 & \underline{48.69} & \underline{36.95} & \underline{67.44} & 88.60 & 26.37 \\
        geometric-only & 53.49 & 46.75 & 30.86 & 43.05 & 79.53 & 22.57 \\
        3D-aware & \underline{55.11} & 48.60 & 36.55 & 67.38 & \underline{92.54} & \underline{26.91} \\
        \midrule
        \rowcolor{lightgray}\textbf{Ours} & \textbf{56.01} & \textbf{49.82} & \textbf{37.59} & \textbf{67.95} & \textbf{96.21} & \textbf{28.36} \\
        \bottomrule
    \end{tabular}
    }
    \label{ablation_merge}
\end{table}

\subsubsection{\textbf{Effect of Hierarchical Cluster-based Pruning.}}
To evaluate the proposed hierarchical pruning strategy, we compare it with token-level pruning without clustering and cluster-level pruning that removes entire clusters according to importance, all under 90\% compression. For fairness, the intra-cluster selection strategy remains identical. As shown in Table~\ref{ablation_prune}, token-level pruning produces fragmented representations and fails to preserve object-level completeness, whereas cluster-level pruning may remove low-importance clusters entirely, reducing spatial coverage. Our hierarchical approach instead allocates budgets across clusters while retaining details within each cluster, thereby preserving both global scene coverage and local detail fidelity under high compression.

\subsubsection{\textbf{Effect of Individual Components.}}
To assess the contributions of the two components, we apply only spatial graph-based token merging (Sec.~\ref{sec:merge}) or hierarchical spatial clustering-based pruning (Sec.~\ref{sec:prune}) under a unified 90\% compression ratio. Table~\ref{tab:ablation_main} shows that both variants underperform the full HiSC framework across ScanRefer, Scan2Cap, and ScanQA. The merging-only variant allocates the entire compression budget before LLM inference and therefore lacks adaptive selection. The pruning-only variant operates on highly redundant tokens, which interferes with budget allocation and weakens important-region representations. These results demonstrate that the two modules are complementary: pre-inference merging safely removes redundancy, while in-LLM hierarchical pruning provides structured and adaptive token allocation.

\begin{figure}[!t]
\centering
\includegraphics[width=0.97\linewidth]{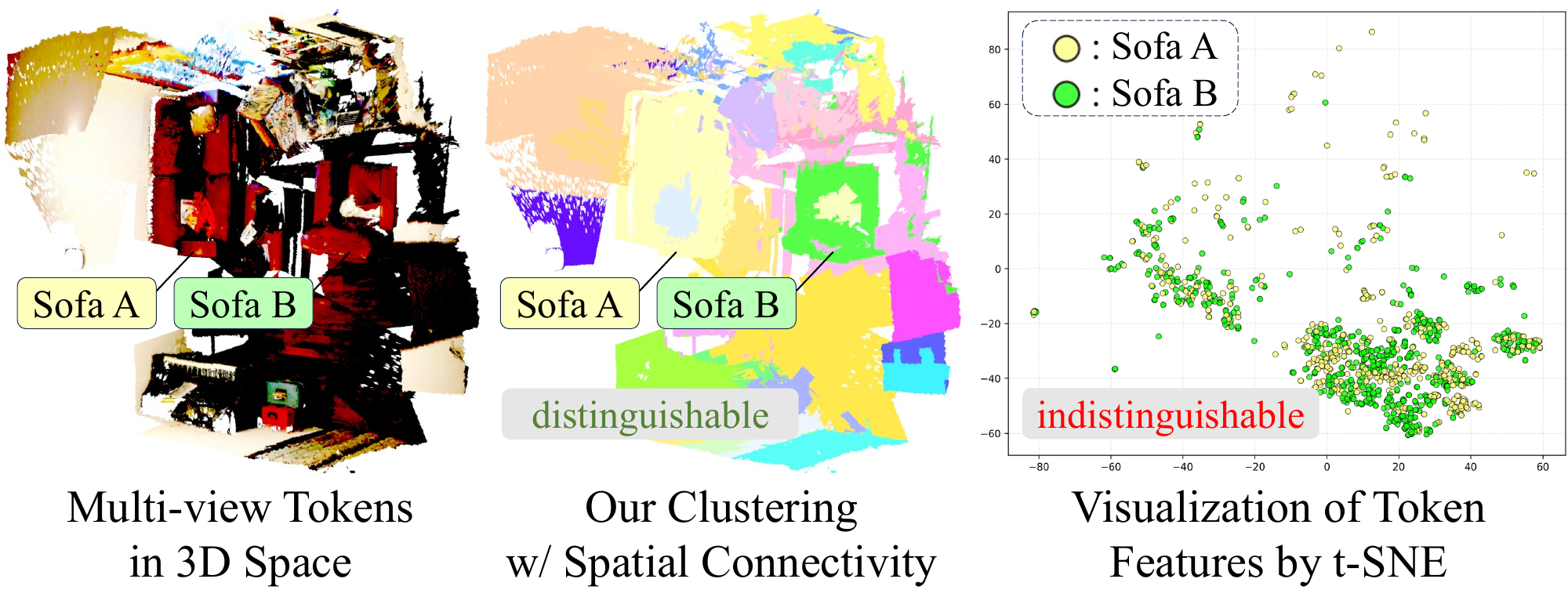}
\caption{Qualitative visualization of connectivity-based clustering. While visually similar objects exhibit overlapping feature distributions, HiSC separates them into distinct clusters by leveraging spatial connectivity, resulting in more accurate object-level grouping.}
\Description{Comparison of two spatially separate sofas. Their
token features overlap in a t-distributed stochastic neighbor
embedding plot, but HiSC uses spatial connectivity to assign the
sofas to distinct clusters.}
\label{fig:merge_vis}
\end{figure}

\subsubsection{\textbf{Qualitative Visualization of Connectivity-based Clustering.}}
Figure~\ref{fig:merge_vis} further illustrates the effect of connectivity-based clustering. Tokens from visually similar but physically distinct objects substantially overlap in feature space, making similarity-based grouping unreliable under appearance ambiguity and cross-view variation. By enforcing spatial connectivity, our method separates such objects while preserving coherent structures within spatially extended instances. The transitivity of connectivity further propagates locally consistent relations across views, yielding more complete and continuous regions. These observations show that 3D connectivity supports reliable object-level grouping for redundancy merging and hierarchical spatial clustering-based pruning.

\begin{table}[t!]
    \centering
    \caption{Comparison of different hierarchical strategies.}
    \renewcommand{\arraystretch}{1.0} 
    \setlength{\tabcolsep}{3pt}
\resizebox{\columnwidth}{!}{
    \begin{tabular}{l|cccccc}
        \toprule 
        \multirow{2}{*}{Setting} & \multicolumn{2}{c}{ScanRefer} & \multicolumn{2}{c}{Scan2Cap}  & \multicolumn{2}{c}{ScanQA} \\ 
         \cmidrule(lr){2-3} \cmidrule(lr){4-5} \cmidrule(lr){6-7}
        & \scalebox{0.9}[1]{Acc@0.25} & \scalebox{0.9}[1]{Acc@0.5} & \scalebox{0.85}[1]{B-4@0.5} & \scalebox{0.85}[1]{C@0.5} & C & EM \\ \midrule
        token-only & \underline{55.16} & \underline{49.05} & \underline{34.38} & \underline{53.32} & \underline{83.63} & \underline{25.09} \\
        cluster-only & 53.49 & 46.75 & 32.56 & 47.06 & 74.84 & 21.90 \\
        \midrule
        \rowcolor{lightgray}\textbf{Ours} & \textbf{56.01} & \textbf{49.82} & \textbf{37.59} & \textbf{67.95} & \textbf{96.21} & \textbf{28.36} \\
        \bottomrule
    \end{tabular}
    }
    \label{ablation_prune}
\end{table}

\begin{table}[t!]
    \centering
    \caption{Contributions of the two proposed components.}
    \renewcommand{\arraystretch}{1.0}
    \setlength{\tabcolsep}{4pt}
\resizebox{\columnwidth}{!}{
    \begin{tabular}{l|cccccc}
        \toprule 
        \multirow{2}{*}{Setting} & \multicolumn{2}{c}{ScanRefer} & \multicolumn{2}{c}{Scan2Cap}  & \multicolumn{2}{c}{ScanQA} \\ 
         \cmidrule(lr){2-3} \cmidrule(lr){4-5} \cmidrule(lr){6-7}
        & \scalebox{0.9}[1]{Acc@.25} & \scalebox{0.9}[1]{Acc@.5} & \scalebox{0.85}[1]{B-4@.5} & \scalebox{0.85}[1]{C@.5} & C & EM \\ \midrule
        w/o SGraM & \underline{55.06} & \underline{48.47} & \underline{36.60} & \underline{67.46} & \underline{91.88} & \underline{27.27} \\
        w/o SCluP & 53.49 & 46.23 & 32.22 & 44.26 & 61.75 & 17.45 \\
        \midrule
        \rowcolor{lightgray}\textbf{Ours} & \textbf{56.01} & \textbf{49.82} & \textbf{37.59} & \textbf{67.95} & \textbf{96.21} & \textbf{28.36} \\
        \bottomrule
    \end{tabular}
    }
    \label{tab:ablation_main}
\end{table}

\begin{figure}[!t]
\centering
\includegraphics[width=0.97\linewidth]{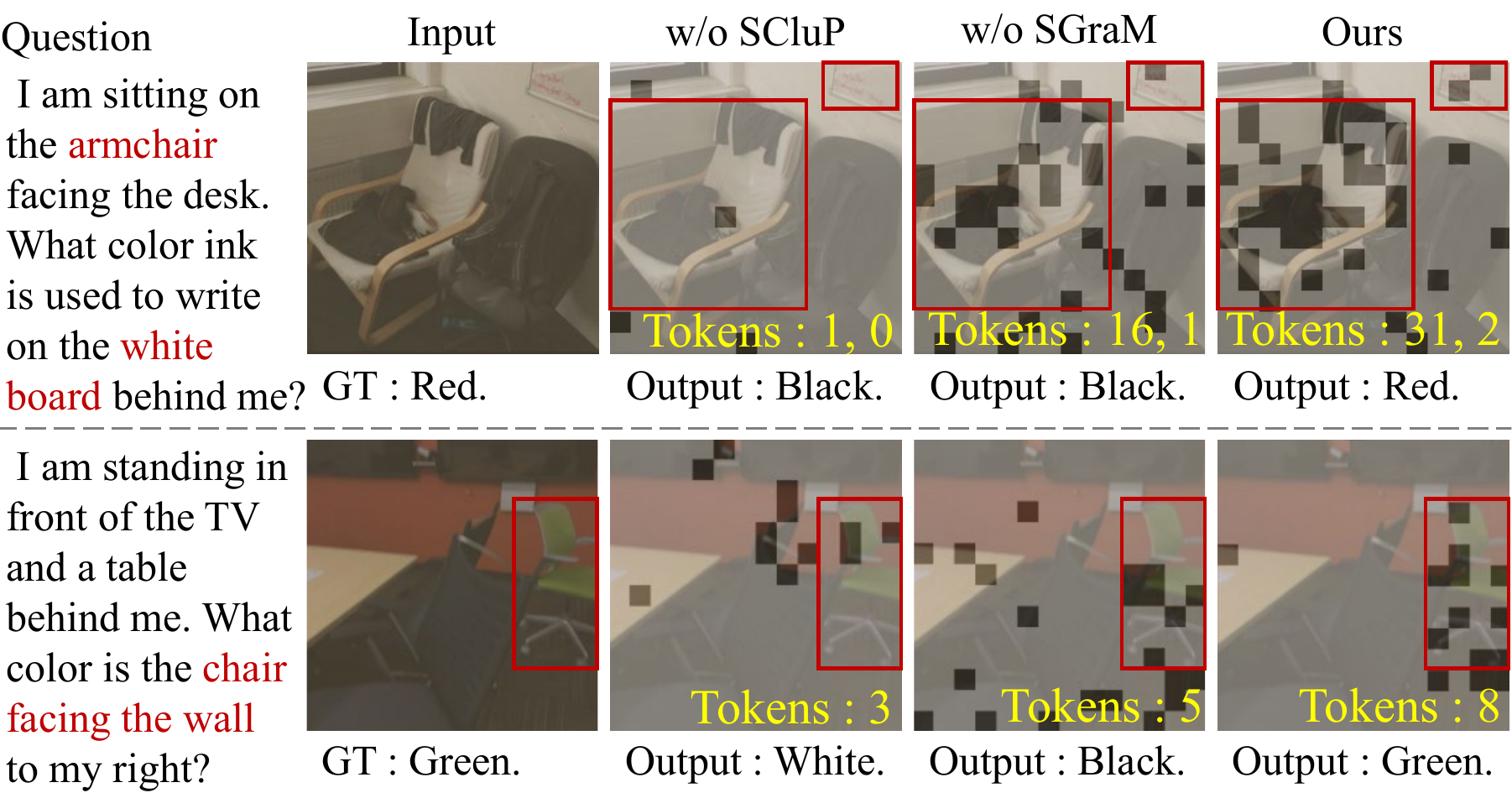}
\caption{Qualitative comparison of pruning results for different components under 90\% compression. For each example, we visualize a representative frame with task-relevant regions (\textcolor{red}{red} boxes) and retained tokens (yellow counts).}
\Description{Two ablation examples under 90 percent token
reduction. Removing spatial clustering-based pruning or spatial
graph-based merging reduces task-relevant token coverage and
produces incorrect color answers, whereas full HiSC retains more
relevant tokens and answers both questions correctly.}
\label{fig:abla_vis}
\end{figure}

\subsubsection{\textbf{Qualitative Visualization of Individual Components.}}
Figure~\ref{fig:abla_vis} visualizes the token distributions produced by the individual components and the full HiSC. With merging alone, redundant regions are effectively combined, but the absence of hierarchical cluster-based pruning can leave important objects with insufficient fine-grained token coverage. Conversely, pruning alone introduces many redundant tokens into the LLM, reducing the effective budget assigned to important regions and causing greater detail loss. HiSC combines both stages: connectivity-based merging first removes redundant tokens before inference, and hierarchical spatial clustering-based pruning then performs structured allocation within the LLM. Together, they balance fine-grained preservation in important regions with complete object-level spatial coverage.

\section{Conclusion}
In this paper, we propose \textbf{HiSC}, a training-free framework for efficient 3D scene understanding via hierarchical spatial clustering-based token compression. Our approach addresses structured redundancy in multi-view 3D inputs, which causes long sequences and inefficient inference. Specifically, we introduce a graph-based geometric-semantic token merging strategy that explicitly models cross-view redundancy as spatial connectivity, reliably merging physically consistent regions while preserving structural integrity. We further propose a hierarchical cluster-based pruning scheme that organizes tokens into object-centric clusters and performs structured token allocation across clusters with adaptive selection within clusters. This design preserves object-level completeness and fine-grained details in important regions throughout the reasoning process, aligning token compression with the hierarchical and spatial characteristics of 3D scenes. 
Experiments across diverse 3D reasoning tasks show that HiSC achieves state-of-the-art performance while significantly improving efficiency, maintaining over 90\% of the original performance under extreme compression and remaining nearly lossless under moderate compression.

\begin{acks}
This work was supported by the National Natural Science Foundation of China (62331006, 625B2026), and the Fundamental Research Funds for the Central Universities.
\end{acks}

\bibliographystyle{ACM-Reference-Format}
\bibliography{main}

\end{document}